\documentclass[sigconf]{acmart}

\copyrightyear{2019}
\acmYear{2019}
\acmConference[WWW '19]{Proceedings of the 2019 World Wide Web Conference}{May 13--17, 2019}{San Francisco, CA, USA}
\acmPrice{}
\acmDOI{10.1145/3308558.3313691}
\acmISBN{978-1-4503-6674-8/19/05}

\usepackage{comment}
\usepackage{multirow}
\usepackage{color}
\usepackage{balance}
\usepackage{algorithm}
\usepackage{algcompatible}
\usepackage{amsmath}
\usepackage{algpseudocode}
\usepackage{float}
\usepackage{amssymb}
\usepackage{fixltx2e}
\definecolor{light-gray}{gray}{0.4}
\usepackage{graphicx} 
\usepackage{subfigure} 
\usepackage{tikz,rotating}
\usetikzlibrary{arrows,shapes,snakes,matrix}
\usepackage{wrapfig,lipsum,booktabs}
\usepackage{enumitem}
\usepackage{arydshln}
\usepackage{framed}
\usepackage{pifont}
\newcommand{\cmark}{\ding{51}}%
\newcommand{\xmark}{\ding{55}}%

\begin{document}
%
\title{Dynamic Deep Multi-modal Fusion for Image Privacy Prediction}

\author{Ashwini Tonge}
\affiliation{Department of Computer Science \\
  \institution{Kansas State University}
}
\email{atonge@ksu.edu}

\author{Cornelia Caragea}
\affiliation{Department of Computer Science \\
  \institution{University of Illinois at Chicago}
}
\email{cornelia@uic.edu}

\newcommand{\mytilde}{\raise.17ex\hbox{$\scriptstyle\mathtt{\sim}$}}
\hyphenation{op-tical net-works semi-conduc-tor}

\definecolor{ao}{rgb}{0.0, 0.5, 0.0}


\begin{abstract}
With millions of images that are shared online on social networking sites, effective methods for image privacy prediction are highly needed. In this paper, we propose an approach for fusing object, scene context, and image tags modalities derived from convolutional neural networks for accurately predicting the privacy of images shared online. Specifically, our approach identifies the set of most competent modalities on the fly, according to each new target image whose privacy has to be predicted. 
The approach considers three stages to predict the privacy of a target image, wherein we first identify the neighborhood images that are visually similar and/or have similar sensitive content as the target image.  Then, we estimate the competence of the modalities based on the neighborhood images. Finally, we fuse the decisions of the most competent modalities and predict the privacy label for the target image. 
Experimental results show that our approach predicts the sensitive (or private) content more accurately than the models trained on individual modalities (object, scene, and tags) and prior privacy prediction works. Also, our approach outperforms strong baselines, that train meta-classifiers to obtain an optimal combination of modalities.
\end{abstract}



\maketitle

\section{Introduction}
Technology today offers innovative ways to share photos with people all around the world,  making online photo sharing an incredibly popular activity for Internet users. These users document daily details about their whereabouts through images and also post pictures of their significant milestones and private events, e.g., family photos and cocktail parties \cite{yannlecun}. Furthermore, smartphones and other mobile devices facilitate the exchange of information in content sharing sites virtually at any time, in any place. 
Although current social networking sites allow users to change their privacy preferences, this is often a cumbersome task for the vast majority of users on the Web, who face difficulties in assigning and managing privacy settings \cite{Lipford:2008}. Even though users change their privacy settings to comply with their personal privacy preference, they often misjudge the private information in images, which fails to enforce their own privacy preferences \cite{OrekondySF17iccv}.
Thus, new privacy concerns 
\cite{Findlaw}
are on the rise and mostly emerge due to users' lack of understanding that semantically rich images may reveal sensitive information \cite{OrekondySF17iccv,overexposed,Squicciarini:2011:AAP:1995966.1996000,Zerr:2012}. 
For example, a seemingly harmless photo of a birthday party may unintentionally reveal sensitive information about a person's location, personal habits, and friends. 
Along these lines, Gross and Acquisti \cite{Gross:2005:IRP:1102199.1102214} analyzed more than 4,000 Carnegie Mellon University students' Facebook profiles and outlined potential threats to privacy. The authors found that users often provide personal information generously on social networking sites, but they rarely change default privacy settings, which could jeopardize their privacy. Employers often perform background checks for their future employees using social networking sites and about $8\%$ of companies have already fired employees due to their inappropriate social media content \cite{citeulike:10040269}. 
A study carried out by the Pew Research center reported that $11\%$ of the users of social networking sites regret the content they posted \cite{PewResearchPriavcy}. 

Motivated by the fact that increasingly online users' privacy is routinely compromised by using social and content sharing applications \cite{Zheleva:2009}, recently, researchers started to explore machine learning and deep learning models to automatically identify private or sensitive content in images 
\cite{Tran:2016:PFD:3015812.3016006,DBLP:conf/aaai/TongeC16,DBLP:conf/aaai/TongeC18,tongemsm18,Squicciarini2014,Zerr:2012,OrekondySF17iccv}. Starting from the premise that the objects and scene contexts present in images impact images' privacy, many of these studies used objects, scenes, and user tags, or their combination (i.e., feature-level or decision-level fusion) 
 to infer adequate privacy classification for online images. 
{\color{blue}
}

\begin{figure*}[t]
\centering
\begin{small}
\begin{framed}
\vspace{-2mm}
\begin{tabular}{@{}cccc@{}}
\hline
\multicolumn{4}{|c|}{{\bf Single modality is correct}}\\
\hline
Image & Tags & \multicolumn{2}{c}{Probabilities} \\ 
&& Base classifiers & Fusion\\
\hdashline


\multirow{2}{*}{{\includegraphics[scale=0.2]{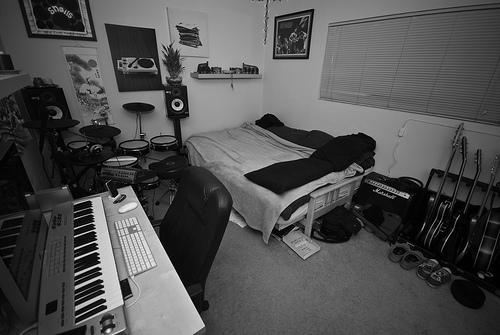}}} & bed, studio & {\bf \color{blue} scene: 0.62} & Feature-level: 0.21\\
& dining table &  object: 0.5 & Decision-level: 0.33\\
(a) & speakers, music & tags: 0.29 & \\

\\


\multirow{2}{*}{{\includegraphics[scale=0.11]{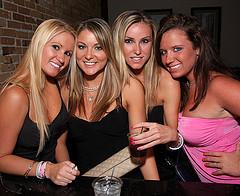}}} & birthday  & scene: 0.57 & Feature-level: 0.21 \\
& night & {\bf \color{blue} object: 0.78} & Decision-level: 0.33 \\ 
(b) & party, life & tags: 0.39 &  \\

\\


\multirow{2}{*}{{\includegraphics[scale=0.07]{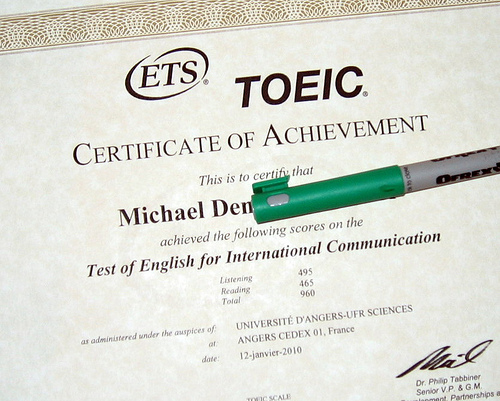}}} & toeic, native  & scene: 0.02 & Feature-level: 0.27 \\
& speaker, text & object: 0.15 & Decision-level: 0.33 \\
(c) & document, pen & {\bf \color{blue} tags: 0.86} & \\



\end{tabular}
\hspace{8mm}
\begin{tabular}{@{}cccc@{}}
\hline
\multicolumn{4}{|c|}{{\bf Multiple modalities are correct}}\\
\hline
Image & Tags & \multicolumn{2}{c}{Probabilities} \\ 
&& Base classifiers & Fusion\\
\hdashline



\multirow{2}{*}{{\includegraphics[scale=0.21]{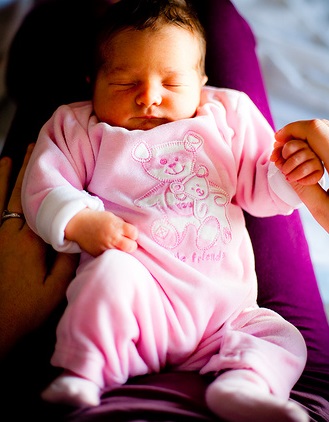}}} & girl, baby  & scene: 0.49 & {\bf \color{blue} Feature-level: 0.77} \\
& indoor, people & {\bf \color{blue} object: 0.87} & {\bf \color{blue} Decision-level: 0.67} \\
(d) & canon & {\bf \color{blue} tags: 0.97} & \\

\\


\multirow{2}{*}{{\includegraphics[scale=0.28]{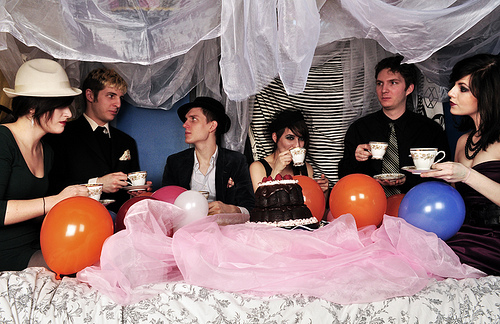}}} & people, party  & {\bf \color{blue} scene: 0.92} & {\bf \color{blue} Feature-level: 0.69} \\
& awesome, tea & object: 0.38 & {\bf \color{blue} Decision-level: 0.67}\\
(e) & bed, blanket & {\bf \color{blue} tags: 0.7} & \\

\\


\multirow{2}{*}{{\includegraphics[scale=0.078]{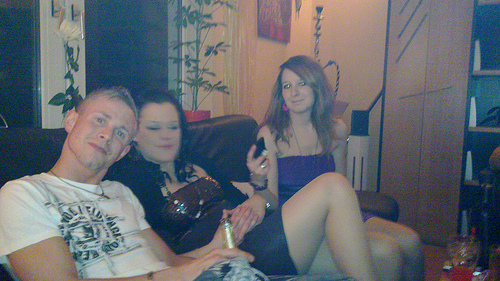}}} & indoor, fun  & {\bf \color{blue} scene: 0.92} & {\bf \color{blue} Feature-level: 0.89}\\
& party & {\bf \color{blue} object: 0.73} & {\bf \color{blue} Decision-level: 1}\\
(f) & people & {\bf \color{blue} tags: 0.77} & \\



\end{tabular}	
\vspace{-1mm}
\end{framed}
\vspace{-4mm}
\end{small}
\caption{Anecdotal evidence of private images and their tags. The feature-level fusion is given as the concatenation of all the features (object, scene, tag) and the decision-level fusion is obtained by averaging the predictions.
} 
\label{fig:motivation}
\end{figure*}

However, we conjecture that simply combining objects, scenes and user tags modalities using feature-level fusion (i.e., concatenation of all object, scene and user tag features) or decision-level fusion (i.e., aggregation of decisions from classifiers trained on objects, scenes and tags) may not always help to identify the sensitive content of images. Figure \ref{fig:motivation} illustrates this phenomenon through several images. For example, let us consider image (a) in the figure. Both feature-level and decision-level fusion models yield very low private class probabilities (Feature-level fusion: $0.21$ and decision-level fusion: $0.33$). Interestingly, a model based on the scene context (bedroom) of the image outputs a high probability of $0.62$, showing that the scene based model is competent to capture the sensitive content of the image on its own. Similarly, for the image (b) (self-portrait) in Figure \ref{fig:motivation}, where scene context is seldom in the visual content, the objects in the image (the ``persons,'' ``cocktail dress'')  are more useful ($0.78$) to predict appropriate image's privacy. Moreover, for images such as ``personal documents'' (image (c)), user-annotated tags provide broader context (such as type and purpose of the document), capturing the sensitive content ($0.86$), that objects and scene obtained through images' content failed to capture. On the other hand, in some cases, we can find more than one competent model for an image (e.g., for image (d)). 
To this end, we propose a novel approach that dynamically fuses multi-modal information of online images, derived through  Convolutional Neural Networks (CNNs), to adequately identify the sensitive image content. In summary, we make the following contributions:

\begin{itemize}
    \item Our significant contribution is to estimate the competence of object, scene and tag modalities for privacy prediction and dynamically identify the most competent modalities for a target image whose privacy has to be predicted.
    \item We derive ``competence'' features from the neighborhood regions of a target image and learn classifiers on them to identify whether a modality is competent to accurately predict the privacy of the target image. To derive these features, we consider privacy and visual neighborhoods of the target image to bring both sensitive and visually similar image content closer.    
    \item We provide an in-depth analysis of our algorithm in an ablation setting, where we record the performance of the proposed approach by removing its various components. The analysis outline the crucial components of our approach.
    \item Our results show that we identify images' sensitive content more accurately than single modality models (object, scene, and tag), multi-modality baselines and prior approaches of privacy prediction, depicting that the approach optimally combines the multi-modality for privacy prediction. 
\end{itemize}

\section{Related Work}
We briefly review the related work as follows.

{\bf Ensemble models and Multi-Modality.}
Several works used ensemble classifiers (or bagging) to improve image classifications \cite{Debeir_texturaland,doi:10.1080/01431160600746456,Ravi:2005:ARA:1620092.1620107}. 
Bagging is an ensemble technique that builds a set of diverse classifiers, each trained on a random sample of the training data to improve the final (aggregated) classifier confidence \cite{Breiman1996,SKURICHINA1998909}. Dynamic ensembles that extend bagging have also been proposed \cite{Dynamiccsel,metades,Cavalin2011DynamicSA} wherein a pool of classifiers are trained on a single feature set (single modality) using the bagging technique \cite{Breiman1996,SKURICHINA1998909}, and the competence of the base classifiers is determined dynamically. 

Ensemble classifiers are also used in the multi-modal setting \cite{Guillaumin,Poria:2016:FAV:2852370.2852645}, where 
different modalities have been coupled, e.g., images and text for image retrieval \cite{pmlr-v32-kiros14} and image classification \cite{Guillaumin}, and audio and visual signals for speech classification \cite{Ngiam:2011:MDL:3104482.3104569}. Zahavy et al. \shortcite{DBLP:conf/aaai/ZahavyKMM18} 
highlighted that classifiers trained on different modalities can vary in their discriminative ability and urged the development of optimal unification methods to combine different classifiers. 
Besides, merging the Convolutional Neural Network (CNN) architectures corresponding to various modalities, that can vary in depth, width, and the optimization algorithm can become very complex. 
However, there is a potential to improve the performance through multi-modal information fusion, which intrigued various researchers \cite{Lynch:2016:IDL:2939672.2939728,pmlr-v32-kiros14,10.1007/978-3-319-10593-2_35,6137235}.  For example, Frome et al. \cite{NIPS2013_5204} merged an image network \cite{NIPS2012_4824} with a Skip-gram Language Model to improve classification on ImageNet. 
Zahavy et al. \shortcite{DBLP:conf/aaai/ZahavyKMM18} proposed a policy network for multi-modal product classification in e-commerce using text and visual content, which learns to choose between the input signals. Feichtenhofer et al. \shortcite{Feichtenhofer16} fused CNNs both spatially and temporally for activity recognition in videos to take advantage of the spatio-temporal information present in videos.  
Wang et al. \shortcite{Object-Scene} designed an architecture to combine object networks and scene networks, which extract useful information such as objects and scene contexts for event understanding. 
Co-training approaches \cite{Blum:1998:CLU:279943.279962} use multiple views (or modalities) to ``guide'' different classifiers in the learning process.  However, co-training methods are semi-supervised and assume that all views are ``sufficient'' for learning. 
%
In contrast with the above approaches, we aim to capture different aspects of images, obtained from multiple modalities (object, scene, and tags), with each modality having a different competence power, 
and perform dynamic multi-modal fusion for image privacy prediction.

{\bf Online Image Privacy.}
 Several works are carried out to study users' privacy concerns in social networks, privacy decisions about sharing resources, and the risk associated with them \cite{Krishnamurthy:2008:CPO:1397735.1397744,Simpson:2008:NUF:1461469.1461470,Ghazinour:2013:MRP:2457317.2457344,DBLP:journals/dke/Parra-ArnauRFME12,6399467,Ilia:2015:FPP:2810103.2813603}. 
\citet{overexposed} examined privacy decisions and considerations in mobile and online photo sharing. They explored critical aspects of privacy such as users' consideration for privacy decisions, content and context based patterns of privacy decisions, how different users adjust their privacy decisions and user behavior towards personal information disclosure. The authors concluded that applications, which could support and influence user's privacy decision-making process should be developed. 
\citet{Jones:2011}  reinforced the role of privacy-relevant image concepts. For instance, they determined that people are more reluctant to share photos capturing social relationships than photos taken for functional purposes; certain settings such as work, bars, concerts cause users to share less.  \citet{1520704} mentioned that users want to regain control over their shared content, but meanwhile, they feel that configuring proper privacy settings for each image is a burden.
Buschek et al. \shortcite{Buschek:2015} presented an approach to assign privacy to shared images using metadata (location, time, shot details) and visual features (faces, colors, edges). Zerr et al. \shortcite{Zerr:2012}  developed the PicAlert dataset, containing Flickr photos, to help detect private images and also proposed a privacy-aware image classification approach to learn classifiers on these Flickr photos. Authors considered image tags and visual features such as color histograms, faces, edge-direction coherence, and SIFT for the privacy classification task. 
Squicciarini et al. \shortcite{Squicciarini2014,Squicciarini:2017:TAO:3062397.2983644} 
 found that SIFT and image tags work best for predicting sensitivity of user's images. 
Given the recent success of CNNs, Tran et al. \shortcite{Tran:2016:PFD:3015812.3016006}, and Tonge and Caragea \shortcite{DBLP:conf/aaai/TongeC16,tongemsm18} showed promising privacy predictions compared with visual features such as SIFT and GIST. 
Yu et al. \shortcite{DBLP:journals/tifs/YuZKLF17} adopted CNNs to achieve semantic image segmentation and also learned object-privacy relatedness to identify privacy-sensitive objects. 

Spyromitros-Xioufis et al. \shortcite{Spyromitros-Xioufis:2016:PPI:2911996.2912018} used features extracted from CNNs to provide personalized image privacy classification, whereas Zhong et  al. \shortcite{zhong2017ijcai} proposed a Group-Based Personalized Model for image privacy classification in online social media sites. 
Despite that an individual's sharing behavior is unique, Zhong et  al. \shortcite{zhong2017ijcai} argued that personalized models generally require large amounts of user data to learn reliable models, and are time and space consuming to train and to store models for each user, while taking into account possible deviations due to sudden changes of users' sharing activities and privacy preferences.  
\citet{OrekondySF17iccv} defined a set of privacy attributes, which were 
first predicted 
from the image content and then used these attributes in combination with users preferences to estimate personalized privacy risk.  The authors used official online social network rules to define the set of attributes, instead of collecting real user's opinions about sensitive content and hence, the definition of sensitive content may not meet a user's actual needs \cite{Li_2018_CVPR_Workshops}.  
Additionally, for privacy attribute prediction, the authors fine-tuned a CNN pre-trained on object dataset. In contrast, we proposed a dynamic multi-modal fusion approach to determine which aspects of images (objects, scenes or tags) are more competent to predict images' privacy. 

\section{Multi-Modality}\label{sec:features}
The sensitive content of an image can be perceived by the presence of one or more objects, the scenes described by the visual content and the description associated with it in the form of tags \cite{tongemsm18,DBLP:conf/aaai/TongeC16,DBLP:conf/aaai/TongeC18,Squicciarini2014}. We derive features (object, scene, tags) corresponding to the multi-modal information of online images as follows. 
{\bf Object ($F^{o}$):}
Detecting objects from images 
is clearly fundamental to assessing whether an image  is of private nature. For example, a single element such as a firearm, political signs, may be a strong indicator of a private image. Hence, we explore the image descriptions extracted from VGG-16 \cite{DBLP:journals/corr/SimonyanZ14a}, a CNN pre-trained on the ImageNet dataset \cite{ILSVRC15} that has $1.2$M+ images labeled with $1,000$ object categories. 
The VGG-16 network implements a $16$ layer deep network; a stack of convolutional layers with a very small receptive field: $3 \times 3$ followed by fully-connected layers. The architecture contains $13$ convolutional layers and $3$ fully-connected layers. 
The input to the network is a fixed-size $224 \times 224$ RGB image. The activation of the fully connected layers capture the complete object contained in the region of interest. Hence,  we use the activation of the last fully-connected layer of VGG-16, i.e., fc$_8$ as a feature vector. The dimension of object features $F^{o}$ is $1000$.

{\bf Scene ($F^{s}$):} 
As consistently shown in various user-centered studies \cite{overexposed}, the context of an image is a potentially strong indicator of what type of message or event users are trying to share online. These scenes, e.g., some nudity, home, fashion events, concerts are also often linked with certain privacy preferences. Similar to object features, we obtain the scene descriptors derived from the last fully-connected layer of the pre-trained VGG-16 \cite{NIPS2012_4824} on the Places2 dataset which contains $365$ scene classes within $2.5$ million images \cite{zhou2016places}. The dimension of scene features $F^{s}$ is $365$.


{\bf Image Tags ($F^{t}$):}
For image tags, we employ 
the CNN architecture of \citet{Collobert:2011:NLP:1953048.2078186}. 
The network contains one convolution layer on top of word vectors obtained from an unsupervised neural language model.  The first layer embeds words into the word vectors pre-trained by \citet{DBLP:conf/icml/LeM14} on 100 billion words of Google News, and are publicly available. The next layer performs convolutions  on the embedded word vectors using multiple filter sizes of $3$, $4$ and $5$, where we use $128$ filters from each size and produce a tag feature representation. A max-pooling operation over the feature map is applied to capture the most important feature of length $256$ for each feature map.  
To derive these features, we consider two types of tags: (1) user tags, and (2) deep tags. 
Because not all images on social networking sites have user tags 
or the set of user tags is very sparse  \cite{mum12}, 
we predict the top $d$ object categories (or deep tags) from the probability distribution extracted from CNN. 

{\bf Object + Scene + Tag ($F^{ost}$):}
We use the combination of the object, scene, and tag features to identify the neighborhood of a target image. 
We explore various ways given in \cite{Feichtenhofer16} to combine the features. For example, we use fc$_7$ layer of VGG to extract features of equal length of $4096$ from both object-net and scene-net and consider the max-pooling of these vectors to combine these features. Note that, in this work, we only describe the combination of features that worked best for the approach. We obtain high-level object ``$F^{o}$'' and scene ``$F^{s}$'' features from fc$_8$ layer of object-net and scene-net respectively and concatenate them with the tag features as follows:
$F^{ost} = f^{cat}(F^{o}, F^{s}, F^{t})$. $F^{ost} = F^{o}(i), 1 \leq i \leq 1000,  F^{ost}(i + 1000) = F^{s}(i), 1\leq i \leq 365, F^{ost}(i + 1365) = F^{t}(i), 1\leq i \leq 256$.

\section{Proposed approach}
We seek to classify a given image into one of the two classes: {\em private} or {\em public}, based on users' general privacy preferences. 
To achieve this, we depart from previous works 
that use the same model 
on all image types (e.g., portraits, bedrooms, and legal documents), 
and propose an approach called ``Dynamic Multi-Modal Fusion for Privacy Prediction'' (or DMFP), that effectively fuses 
multi-modalities (object, scene, and tags) 
and dynamically captures different aspects or particularities from image. Specifically, the proposed approach aims to estimate the competence of models 
trained on these individual modalities for each target image (whose privacy has to be predicted) 
and dynamically identifies the subset of the most ``competent'' models for that image. 
The rationale for the proposed method is that for a particular type of sensitive content, some modalities may be important, whereas others may be irrelevant and may simply introduce noise. 
Instead, a smaller subset of modalities may be significant 
in capturing a particular type of sensitive content (e.g., objects for portraits, scenes for interior home or bedroom, and tags for legal documents, as shown in Figure \ref{fig:motivation}).  

The proposed approach considers three stages to predict the privacy of a target image, wherein we first identify the neighborhood images that are visually similar and/or have similar sensitive content as the target image (Section \ref{sec:neighborhood}).  Then, using a set of base classifiers, each trained on an individual modality, we estimate the competence of the modalities by determining which modalities classify the neighborhood images correctly (Section \ref{sec:meta}). The goal here is to select the most competent modalities for a particular type of images (e.g., scene for home images). Finally, we fuse the decisions of the most competent base classifiers (corresponding to the most competent modalities) and predict the privacy label for the target image (Section \ref{sec:fusion}).

Our approach considers two datasets, denoted as $\mathcal{D}^T$ and $\mathcal{D}^E$, that contain images labeled as {\em private} or {\em public}. We use the dataset $\mathcal{D}^T$ to train a base classifier for each modality to predict whether an image is public or private. Particularly, we train 3 base classifiers $\mathcal{B} = \{B^{o}, B^{s}, B^{t}\}$ on the corresponding modality feature sets from $\mathcal{F}$. 
Note that we use the combination of feature sets $F^{ost}$ only for visual content similarity and do not train a base classifier on it.
The competences of these base classifiers are estimated on the $\mathcal{D}^E$ dataset. We explain the stages of the proposed approach as follows. The notation used is shown in Table \ref{table:notations}.

\begin{table}[h]
\centering
\begin{small}
\begin{tabular}{|p{0.7cm}p{7.1cm}|}
\hline
Notation & \multicolumn{1}{c|}{Description} \\
$\mathcal{D}^T$ & $ = \{(I_1, Y_1),\cdots,(I_m, Y_m)\}$ a dataset of labeled images for base classifier training.\\
$\mathcal{D}^E$ & $ = \{(X_1, Y_1),\cdots,(X_n, Y_n)\}$ a dataset of labeled images for competence estimation.\\
$T$ & A target image with an unknown privacy label.\\
$\mathcal{F}$ & $ = \{F^{o}, F^{s}, F^{t}, F^{ost}\}$ a collection of modality feature sets of object, scene, tag, and object+scene+tag, respectively.\\
$\mathcal{B}$ & $ = \{B^{o}, B^{s}, B^{t}\}$ a set of base classifiers trained on corresponding modality feature sets from $\mathcal{F}$ (e.g., $B^{o}$ is trained on $F^{o}$).\\ 
${N}^T_V$ & The visual similarity based neighborhood of image $T$ estimated using visual content features $F^{ost}$, i.e., the set of most similar images to $T$ based on visual content. \\
$k_v$ & The size of ${N}^T_V$, where $1 \leq k_v < n$. \\ 
${N}^T_P$ & The privacy profile based neighborhood of target $T$, i.e., the set of most similar images to $T$ based on images' privacy profiles. \\
$k_{p}$ & The size of ${N}^T_P$, where $1 \leq k_{p} < n$. \\  
$C$ & $ = \{C^{o}, C^{s}, C^{t}\}$ a set of ``competence'' classifiers corresponding to the base classifiers from $\mathcal{B}$ (e.g., $C^{o}$ for $B^{o}$).\\
$\Phi$ & $ = \{\phi^o, \phi^s, \phi^t\}$ a set of ``competence'' feature vectors for training the ``competence'' classifiers.\\
\hline
\end{tabular}
\end{small}
\caption{Mathematical notations.} 
\label{table:notations}
\vspace{-8mm}
\end{table}

\subsection{Identification of Neighborhoods}\label{sec:neighborhood}
The competence of a base classifier is estimated based on a local region where the target image is located. Thus, given a target image $T$, we first estimate two neighborhoods for $T$: (1) visual similarity based (${N}^T_V$) and (2) privacy profile based (${N}^T_P$) neighborhoods.

The neighborhood ${N}^T_V$ of target image $T$ consists of the $k_v$ most similar images from $\mathcal{D}^E$ using visual content similarity. Specifically, using the $F^{ost}$ features obtained by concatenating object, scene, and tag features (as explained in Section \ref{sec:features}), we determine the $k_v$ most visually similar images to $T$ by applying the K-Nearest Neighbors algorithm on the $\mathcal{D}^E$ dataset.

The neighborhood ${N}^T_P$ of target image $T$ consists of $k_{p}$ most similar images to $T$ by calculating the cosine similarity between the privacy profile of $T$ and images from the dataset $\mathcal{D}^E$. We define privacy profile (denoted by $\overline{T}$) of image $T$ as a vector of posterior privacy probabilities obtained by the base classifiers $\mathcal{B}$ i.e., $\overline{T} = \bigcup_{B_i \in \mathcal{B}} \{P(Y_T = private | T,B_i), P(Y_T = public | T,B_i)\}$. For image (a) in Figure \ref{fig:motivation}, $\overline{T} = [0.62, 0.38, 0.5, 0.5, 0.29, 0.71]$. We consider the privacy profile of images because images of particular image content (bedroom images or legal documents) tend to possess similar privacy probabilities with respect to the set of base classifiers $\mathcal{B}$. For instance, irrespective of various kinds of bedroom images, the probabilities for a private class obtained by base classifiers $\mathcal{B}$, would be similar. This enables us to bring sensitive content closer irrespective of their disparate visual content. Moreover, we consider two different numbers of nearest neighbors $k_v$ and $k_{p}$ to find the neighborhoods since the competence of a base classifier is dependent on the neighborhood and estimating an appropriate number of neighbors for the respective neighborhoods reduces the noise.

\subsection{``Competence'' Estimation}\label{sec:meta}

We now describe how we estimate the ``competence'' of a base classifier. For instance, for the image (a) in Figure \ref{fig:motivation}, scene model has a higher competence than the others, and here, we capture this competence through ``competence'' features and ``competence'' classifiers. 
Specifically, we train a competence classifier for each base classifier that predicts if the base classifier is competent or not for a target image $T$. The features for learning the competence classifiers and the competence learning are described below. 
\subsubsection{Derivation of ``Competence'' Features}

\begin{figure*}[t]
\centering

 \tikzstyle{rectangle}=[draw=gray,thick]
 \tikzstyle{entity}=[fill=gray!25,draw=gray!50,thick,text width=0.54cm, font=\footnotesize, text centered] 
  \tikzstyle{entity2}=[fill=gray!25,draw=gray!50,thick,text width=1cm, font=\footnotesize, text centered] 
    \tikzstyle{entity4}=[fill=gray!25,draw=gray!50,thick,text width=1.45cm, font=\footnotesize, text centered] 
    \tikzstyle{entity5}=[draw=gray!50,thick,text width=2.45cm, font=\footnotesize, text centered] 
    \tikzstyle{entity3}=[draw=black,fill=gray!25,semithick,text width=0.12cm, text height=0.12cm, font=\footnotesize, text centered] 
\tikzstyle{cut}=[circle,draw=black,fill=gray!70,semithick, inner sep=0pt,minimum size=2mm]
\tikzstyle{node}=[circle,draw=black,semithick,inner sep=0pt,minimum size=2mm] 
\tikzstyle{cut}=[circle,draw=black,fill=gray!50,thin, inner sep=0pt,minimum size=4mm]   
\tikzstyle{node3}=[circle,draw=gray!80,thick,inner sep=0pt,minimum size=3mm] 
\tikzstyle{node4}=[circle,draw=black,thick,inner sep=0pt,minimum size=3mm] 
\tikzstyle{node5}=[circle,draw=magenta!80!black,thick,inner sep=0pt,minimum size=3mm] 
\tikzstyle{db}=[cylinder,draw=black,shape border rotate=90, minimum height=60, minimum width=70, outer sep=-0.5\pgflinewidth]

\begin{tikzpicture}

\draw[draw=gray] (1.0,7.0) -- (1.0,15) -- (18.8,15)-- (18.8,7.0) -- (1.0,7.0);

\node at (2.8,14.7) {\large {Target image $T$ ({\color{ao}Private})}};

\node at (2.7,13.7) {\includegraphics[scale=0.5]{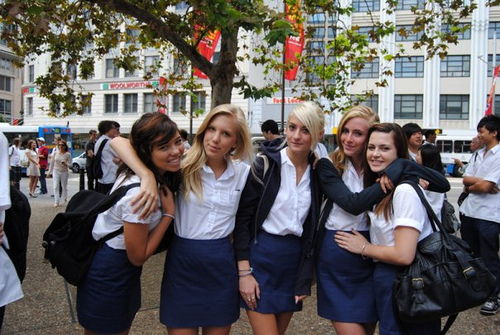}};


\node at (2.7,12.7) {Presentation, Day, School};
\node at (2.7,12.35) {Town-hall, Girls, People};
\node at (2.7,12.1) {Outdoor};

\draw[->, line width=1mm] (3,11.5) to (3,11) to (3.5,11);

\node at (9.5,14.7) {\Large {\#1 Neighborhoods}};
\draw[draw=gray, dash dot] (5.2,9.9) -- (5.2,14.45) -- (13.8,14.45)-- (13.8,9.9) -- (5.2,9.9);

\node at (7.5,14.2) {\large {${N}^T_V$ ($k_v = 7$)}};
\node at (11.5,14.2) {\large {${N}^T_P$ ($k_p = 5$)}};
\node at (9.5,12) {\includegraphics[scale=0.47]{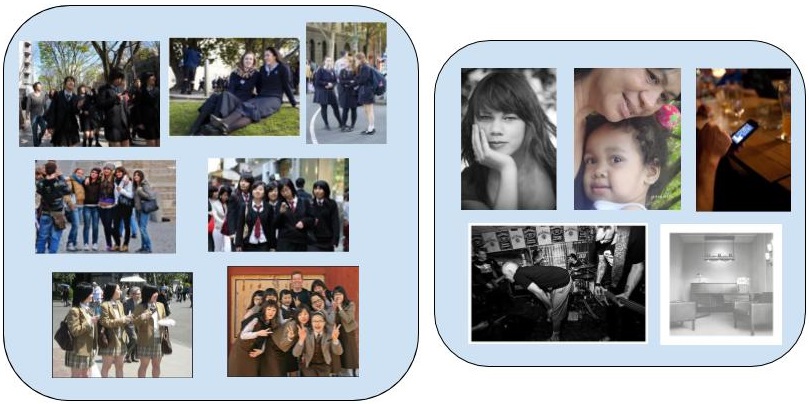}};

\draw[->, line width=1mm] (6.5,9.7) to (6.5,9.3);

\node at (9.5,9.6) {\Large {\#2 ``Competence'' Features ($\Phi$)}};

\node at (5.45,9.1) {$\phi^o$};
\matrix[matrix of nodes, row sep=3\pgflinewidth, column sep=3\pgflinewidth,
               nodes={rectangle, draw=gray, minimum height=1.2em, minimum width=1.2em,
                      anchor=center, 
                      inner sep=2pt, outer sep=0pt}] at (9.0,9.0) 
{ {1} && {1} && {0} && {1} &&
  {0} && {1} && {1} && {1} &&
  {1} && {1} && {1} && {1} && {0.67}
\\ };

\draw [decorate,decoration={brace,amplitude=8pt,mirror,raise=4pt},yshift=0pt]
(5.7,9.0) -- (9.05,9.0) node [black,midway,yshift=-0.6cm] {$\phi_1$};



\draw [decorate,decoration={brace,amplitude=8pt,mirror,raise=4pt},yshift=0pt]
(9.1,9.0) -- (11.55,9.0) node [black,midway,yshift=-0.6cm] {$\phi_2$};

\draw [decorate,decoration={brace,amplitude=7pt,mirror,raise=4pt},yshift=0pt]
(11.6,9.0) -- (12.3,9.0) node [black,midway,yshift=-0.6cm] {$\phi_3$};


\node at (5.45,8.0) {$\phi^s$};
\matrix[matrix of nodes, row sep=3\pgflinewidth, column sep=3\pgflinewidth,
               nodes={rectangle, draw=gray, minimum height=1.2em, minimum width=1.2em,
                      anchor=center, 
                      inner sep=2pt, outer sep=0pt}] at (9.0,8.0) 
{ {1} && {0} && {1} && {1} &&
  {1} && {1} && {1} && {0} &&
  {0} && {0} && {0} && {0} && {0.58}
\\ };



\node at (5.45,7.5) {$\phi^{t}$};
\matrix[matrix of nodes, row sep=3\pgflinewidth, column sep=3\pgflinewidth,
               nodes={rectangle, draw=gray, minimum height=1.2em, minimum width=1.2em,
                      anchor=center, 
                      inner sep=2pt, outer sep=0pt}] at (9.0,7.5) 
{ {0} && {0} && {0} && {1} &&
  {0} && {1} && {1} && {1} &&
  {1} && {1} && {1} && {1} && {0.99}
\\ };

\draw[draw=gray, dash dot] (5.2,9.3) -- (5.2,7.2) -- (12.45,7.2)-- (12.45,9.3) -- (5.2,9.3);


\draw[draw=gray, dash dot] (14.1,7.2) -- (14.1,14) -- (18.66,14)-- (18.66,7.2) -- (14.1,7.2);





\draw[->, line width=1mm] (13,8) to (13.5,8);

\node at (16.3,14.7) {\Large {\#3 Dynamic Fusion of}};
\node at (16.5,14.3) {\Large {Multi-Modality}};

\node at (16.4,11.4) {\includegraphics[scale=0.475]{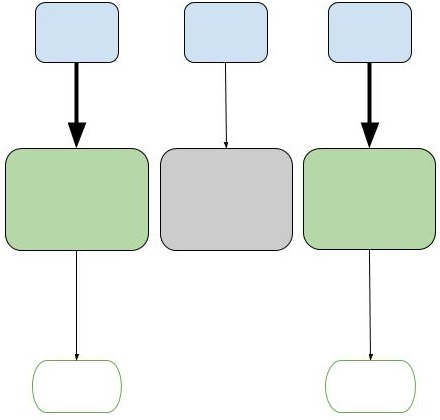}};

\node at (16.3,13.8) {Is a base classifier competent?};
\node at (16.35,10.5) {Competent base classifiers};
\node at (16.35,10.2) {are selected.};

\node at (15,13.15) {$C^o$};
\node at (16.5,13.15) {$C^s$};
\node at (17.9,13.15) {$C^t$};

\node at (15,11.5) {$B^o$};
\node at (14.95,11.75) {\footnotesize{$CS_o = 0.97$}};
\node at (14.95,11.21) {\footnotesize{Object CNN}};
\node at (16.5,11.5) {$B^s$};
\node at (16.45,11.75) {\footnotesize{$CS_s = 0.08$}};
\node at (16.45,11.21) {\footnotesize{Scene CNN}};
\node at (17.9,11.5) {$B^t$};
\node at (17.85,11.75) {\footnotesize{$CS_t = 0.99$}};
\node at (17.85,11.21) {\footnotesize{Tag CNN}};

\node at (15,9.65) {\footnotesize{\color{ao} Private}};
\node at (17.9,9.65) {\footnotesize{\color{ao} Private}};

\node at (14.9,9.15) {\#Votes};

\node at (16.5,8.7) {Private : $0.97$ $(CS_o) + 0.99$ $(CS_t)$};
\node at (16,8.4) {$ = 1.96$};
\node at (15.1,8.1) {Public : 0};

\draw[draw=gray] (14.35,7.9) -- (14.35,9.2) -- (14.4,9.2);

\draw[draw=gray] (15.5,9.2) -- (18.6,9.2)-- (18.6,7.9) -- (14.35,7.9);

\node at (16.6,7.5) {\large{Majority Vote: {\color{ao}Private}}};

\end{tikzpicture} 
\vspace{-8mm}
\caption{Illustration of the proposed approach using an example.}
\vspace{-2mm}
\label{fig:AlgIll}
\end{figure*}

 We define three different sets of ``competence'' features wherein each set of these features captures a different criterion to estimate the level of competence of base classifiers dynamically. 
 The first competence feature $\phi_1$, for image $T$, is derived from the neighborhood ${N}^T_V$ (based on visual similarity) whereas the second competence feature $\phi_2$ is obtained from the neighborhood ${N}^T_P$ (based on privacy profile).
 The third competence feature $\phi_3$ captures the level of confidence of base classifiers for predicting the privacy of the image ($T$) itself. We create a ``competence'' feature vector by concatenating all these competence features $\phi = \{ \phi_1 \cup \phi_2 \cup \phi_3 \}$  into a vector of length $|\phi| = k_v + k_{p} + 1$. We extract such competence vectors corresponding to each base classifier in $\mathcal{B}$ (e.g., $\phi^o$ for $B^o$, refer to Figure \ref{fig:AlgIll}). We extract these ``competence'' features as follows.

{$\phi_1$}: A vector of $k_v$ entries that captures the correctness of a base classifier in the visual neighborhood region $N_V^T$. 
An entry $j$ in $\phi_1$ is 1 if a base classifier $B_i \in \mathcal{B}$ accurately predicts privacy of image $X_j \in N_V^T$, 
and is 0 otherwise, where $j = 1,\cdots,k_v$. For the target image in Figure \ref{fig:AlgIll}, $\phi_1 = \{1, 1, 0, 1, 0, 1, 1\},$ obtained by $B^o$.



{$\phi_2$}: A vector of $k_p$ entries that captures the correctness of a base classifier in the privacy profile neighborhood region $N_P^T$. 
An entry $j$ in $\phi_2$ is 1 if a base classifier $B_i \in \mathcal{B}$ accurately predicts privacy of image $X_j \in N_P^T$, 
and is 0 otherwise, where $j = 1,\cdots,k_p$. For the target image in Figure \ref{fig:AlgIll}, $\phi_2 = \{1, 1, 1, 1, 1\},$ obtained using $B^o$.


{$\phi_3$}: We capture a degree of confidence of base classifiers for target image $T$. Particularly, we consider the maximum posterior probability obtained for target image $T$ using base classifier $B_i$ i.e. $Max(P(Y_T = Private | T,B_i), P(Y_T = Public | T,B_i))$, where $B_i \in \mathcal{B}$. For the target image in Figure \ref{fig:AlgIll}, $\phi_3 = 0.67,$ obtained using $B^o$.

\begin{algorithm}[t]
\caption{The ``Competence'' Learning}
\label{metatraining}
\begin{algorithmic}[1]
\State \textbf{Input}: A dataset $\mathcal{D}^E =\{(X_1, Y_1),\cdots,(X_n, Y_n)\}$ of labeled images; $F^{ost}_{X_j}$ combination of modality feature sets for $X_j$;
a set of base classifiers $\mathcal{B} = \{B^{o}, B^{s}, B^{t}\}.$ 
 
\State \textbf{Output}: A set of ``competence'' classifiers $\mathcal{C}=\{C^{o}, C^{s}, C^{t}\}$.

\State $\mathcal{D}=\{{D}^o, {D}^s, {D}^t\} \gets \varnothing$; {\color{light-gray}{\small{// Datasets for training competence classifiers, initially empty.}}}
\State $\mathcal{C} \gets \varnothing$; {\color{light-gray}{\small{// A set of competence classifiers, initially empty.}}}

\ForAll {$X_j \in \mathcal{D}^E$} 
        	\State $N_V^{X_j} \gets { IdentifyVisualNeighborhood(k_v, X_j, F^{ost}_{X_j}, \mathcal{D}^E)}$; {\color{light-gray}{\small{// $k_v$ nearest neighbors of $X_j$ obtained using visual content similarity.}}} 
            \State $\overline{X_j} \gets { ComputePrivacyProfile(X_j, \mathcal{B})}$; {\color{light-gray}{\small{//Privacy profile.}}} 
            \State $N_P^{X_j} = { IdentifyPrivacyNeighborhood(k_{p}, \overline{X_j}, \overline{\mathcal{D}^E}})$; {\color{light-gray}{\small{// $k_{p}$ most similar images of $\overline{X_j}$ obtained using privacy profile similarity.}}}
              \ForAll {$B_i \in \mathcal{B}$} {\color{light-gray}{\small{// Iterate through the set of base classifiers.}}}
             	\State $\phi_{i,j} \gets { CompetenceFeatures(X_j, N_V^{X_j}, N_P^{X_j}, B_i)}$;
                \If {${ Predict(B_i, X_j)} = Y_j$} {\color{light-gray}{\small{// predicted correctly.}}}
                	\State 	$L_{i,j} \gets 1$; {\color{light-gray}{\small{// $B_i$ is competent for $X_j$.}}} 
                \Else
                	\State 	$L_{i,j} \gets 0$; {\color{light-gray}{\small{// $B_i$ is not competent for $X_j$.}}}          
                \EndIf
		        \State ${D^i} \gets {D^i} \cup \{(\phi_{i,j}, L_{i,j})\}$
                \EndFor
\EndFor
\ForAll {$D^i \in \mathcal{D}$} {\color{light-gray}{\small{// Train competence classifiers.}}} 
    \State $C_i \gets TrainCompetenceClassifier(D^i)$; 
    \State $\mathcal{C} \gets \mathcal{C} \cup C_i$
\EndFor

\State \Return $\mathcal{C}$;

\end{algorithmic}
\end{algorithm}

\begin{algorithm}[t]
\caption{Dynamic Fusion of Multi-Modality}
\label{classifierselection}
\centering
\begin{algorithmic}[1]
\State \textbf{Input}: A target image $T$; $\mathcal{D}^E =\{(X_1, Y_1), \cdots,(X_n, Y_n)\}$ a dataset of labeled images; $F^{ost}_T$ combination of modality feature sets for $T$;
a set of base classifiers $\mathcal{B} = \{B^{o}, B^{s}, B^{t}\}$; and a set of competence classifiers $\mathcal{C} = \{C^{o}, C^{s}, C^{t}\}.$ 
 
\State \textbf{Output}:  Privacy label $Y_T$.

\State $B' \gets \varnothing$; {\color{light-gray}{\small{// the subset of most competent base classifiers.}}}  
\State $CS \gets \varnothing$; {\color{light-gray}{\small{// the set of competence scores.}}} 
\State $T_{ba} \gets {Agreement(\mathcal{B}, T)}$; {\color{light-gray}{\small{// Base classifiers' agreement on $T$'s label.}}}
\If{$T_{ba} \leq |\mathcal{B}|$}
	\State $N_V^{T} \gets { IdentifyVisualNeighborhood(k_v, T, F^{ost}_T, \mathcal{D}^E)}$; {\color{light-gray}{\small{// $k_v$ nearest neighbors of $T$ obtained using visual content similarity.}}}
	
    \State $\overline{T} \gets { ComputePrivacyProfile(T, \mathcal{B})}$; {\color{light-gray}{\small{//Privacy profile of $T$.}}} 
    \State $N_P^{T} = { IdentifyPrivacyNeighborhood(k_{p}, \overline{T}, \overline{\mathcal{D}^E}})$; {\color{light-gray}{\small{// $k_{p}$ most similar images of $\overline{T}$ obtained using privacy profile similarity.}}}	
	
    \ForAll {$B_i \in \mathcal{B} \And C_i \in \mathcal{C}$} {\color{light-gray}{\small{// Iterate through the set of base and competence classifiers.}}} 
 	    \State $\phi_i \gets {CompetenceFeatures(T, N_V^{T}, N_P^{T}, B_i)}$;    
        \State 	$CS_i \gets PredictCompetence(F_i, C_i)$; {\color{light-gray}{\small{// Predict competence score for base classifier $B_i$.}}}       
        \If{$CS_i > 0.5$} {\color{light-gray}{\small{// If the predicted competence score is greater than 0.5 then the base classifier $B_i$ is competent.}}}

           	\State $B' \gets B' \cup \{B_{i}\}$ 
           	\State $CS \gets CS \cup \{CS_{i}\}$
         \EndIf
    \EndFor
    
   \State $Y_{T} = WeightedMajorityVote(T, B', CS)$ {\color{light-gray}{\small{// Votes are first weighted by the competence score and then majority vote is taken.}}}
 \EndIf

\State \Return $Y_{T}$
\end{algorithmic}
\end{algorithm}

\subsubsection{``Competence'' Learning}

We learn the ``competence'' of a base classifier by training a binary ``competence'' classifier on the dataset $\mathcal{D}^E$ in a Training Phase. A competence classifier predicts whether a base classifier is competent or not for a target image. Algorithm \ref{metatraining} describes the ``competence'' learning process in details. Mainly, we consider images from $\mathcal{D}^E$  as target images (for the training purpose only) and identify both the neighborhoods (${N_V}$,  ${N_P}$) from the dataset $\mathcal{D}^E$ itself (Alg. \ref{metatraining}, lines 6--8). Then, we extract ``competence'' features for each base classifier in $\mathcal{B}$ based on the images from these neighborhoods (Alg. \ref{metatraining}, line 10). To reduce noise, we extract ``competence'' features by considering only the images belonging to both the neighborhoods. On these ``competence'' features, we train a collection of ``competence'' classifiers $\mathcal{C}$ corresponding to each base classifier in $\mathcal{B}$ (Alg. \ref{metatraining}, lines 19--22). Precisely, we train $3$ competence classifiers $\mathcal{C} = \{C^o, C^s, C^t\}$. To train ``competence'' classifier $C_i \in \mathcal{C}$, we consider label $L_i = 1$ if base classifier $B_i \in \mathcal{B}$ predicts the correct privacy of a target image (here, $X_j \in \mathcal{D}^E$), otherwise 0  (Alg. \ref{metatraining}, lines 11--16). 

\subsection{Dynamic Fusion of Multi-Modality}\label{sec:fusion}
In this stage, for given target image $T$, we dynamically determine the subset of most competent base classifiers. We formalize the process of base classifier selection in Algorithm \ref{classifierselection}. The algorithm first checks the agreement on the privacy label between all the base classifiers in $\mathcal{B}$ (Alg. \ref{classifierselection}, line 5). If not all the base classifiers agree, then we estimate the competence of all the base classifiers and identify the subset of most competent base classifiers for the target image as follows. Given target image $T$, Algorithm \ref{classifierselection} first identifies both the neighborhoods (${N}^T_V$,  ${N}^T_P$) using the visual features $F^{ost}$ and privacy profile from $\mathcal{D}^E$ dataset (Alg. \ref{classifierselection}, lines 7--9). Using these neighborhoods, we extract ``competence'' feature vectors (explained in Section \ref{sec:meta}) and provide them to the respective ``competence'' classifiers in $\mathcal{C}$ (learned in the Training Phase) to predict competence score of base classifier $B_i$. If the competence score is greater than $0.5$, then base classifier $B_i$ is identified as competent to predict the privacy of target image $T$ (Alg. \ref{classifierselection}, lines 10--17). Finally, we weight votes of the privacy labels predicted by the subset of most competent base classifiers by their respective ``competence'' score and take a majority vote to obtain the final privacy label for target image $T$ (Alg. \ref{classifierselection}, line 18). A ``competence'' score $CS_i$ is given as a probability of base classifier $B_i$ being competent. We consider the majority vote of the most competent base classifiers because certain images (e.g., vacation) might require more than one base classifiers (object and scene) to predict the appropriate privacy. If both the privacy classes (private and public) get the same number of votes, then the class of a highest posterior probability is selected.

\subsubsection*{\bf Illustration of the Proposed Approach }
Figure \ref{fig:AlgIll} shows the illustration of the proposed approach through an anecdotal example. We consider a target image $T$ whose privacy has to be predicted. For $T$, we first identify two neighborhoods: (1) visual content ($N_V^T$), 2. privacy profile ($N_P^T$). For $N_V^T$, we use visual content features $F^{ost}$ to compute the similarity between target image $T$ and the images from the dataset $\mathcal{D}^E$. The top $k_v = 7$ similar images for $N_V^T$ are shown in the figure (left blue rectangle).  Likewise, for $N_P^T$, we compute the similarity between privacy profile of the target image $\overline{T}$ and privacy profiles of images in $\mathcal{D}^E$. We show the top $k_p = 5$ similar images for $N_P^T$ in the right blue rectangle of Figure \ref{fig:AlgIll}. 
From these neighborhoods, we derive a ``competence'' feature vector $\phi$ for each base classifier in $\mathcal{B}$ (e.g., $\phi^o$ for $B^o$). We show these ``competence'' features in the figure as a matrix of feature values. We input these features to the respective ``competence'' classifiers from $\mathcal{C}$ (e.g., $\phi^o$ to $C^o$), that predict whether a base classifier $B_i \in B$ is competent to predict correct privacy label of the target image ($T$). The ``competence'' classifiers ($C^o, C^s, C^t$) are shown as blue rectangles on the right side of Figure \ref{fig:AlgIll}. The base classifiers $B^{o}$ and $B^{t}$ are predicted as competent and hence are selected to obtain the final privacy label for the target image. The competent base classifiers are shown in green rectangles on the right side of Figure \ref{fig:AlgIll}. Once we select the competent base classifiers, we take a weighted majority vote on the privacy labels, predicted by these base classifiers. For example, in this case, the competent base classifiers $B^{o}$ and $B^{t}$ predict the privacy of $T$ as ``private,'' and hence, the final privacy label of $T$ is selected as ``private.'' It is interesting to note that the target image ($T$) contains ``outdoor'' scene context that is not useful to predict the correct privacy label and hence, the scene model  $B^{s}$ is not selected by the proposed approach for the target image.

\section{Dataset}
We evaluate our approach on a subset of $32,000$ Flickr images sampled from the PicAlert dataset, made available by Zerr et al. \shortcite{Zerr:2012}. PicAlert consists of Flickr images on various subjects, which are manually labeled as {\em public} or {\em private} by external viewers. 
The guideline to select the label is given as: private images belong to the private sphere (like self-portraits, family, friends, someone's home) or contain information that one would not share with everyone else (such as private documents). The remaining images are labeled as public. 
The dataset of $32,000$ images is split in $\mathcal{D}^T$, $\mathcal{D}^E$ and {\em Test} sets of $15,000$, $10,000$ and $7,000$ images, respectively. Each experiment is repeated $5$ times with a different split of the three subsets (obtained using $5$ different random seeds) and the results are averaged across the five runs. 
The public and private images are in the ratio of 3:1 in all subsets. 

\section{Experiments and Results}

We evaluate the privacy prediction performance obtained using the proposed approach DMFP, where we train a set of base classifiers $\mathcal{B}$ on images in the dataset $\mathcal{D}^T$, and dynamically estimate the ``competence'' of these base classifiers for target images in {\em Test} by identifying neighborhoods (${N_V}, {N_P}$) using images in $\mathcal{D}^E$. We first consider various values of neighborhood parameters $k_v$ and $k_{p}$ and show their impact on the performance of the proposed approach. Then, we compare the performance of the proposed approach with the performance obtained using three types of mechanisms: 
(1) components of the proposed approach, that are used to fuse the multi--modal characteristics of online images, (2) the state-of-the-art approaches for privacy prediction, and (3) strong baselines that select models based on their competence (e.g., \citet{DBLP:conf/aaai/ZahavyKMM18}) and that attempt to yield the optimal combination of base classifiers (for instance, using stacked ensemble classifiers).


{\bf \em Evaluation Setting.} 
We train base classifiers ($\mathcal{B}$) using the Calibrated linear Support Vector Machine (SVM) implemented in Scikit-learn library 
to predict more accurate probabilities. We use 3-fold cross-validation on the dataset $\mathcal{D}^T$ to fit the linear SVM on the $2-$folds, and the remaining fold is used for calibration. The probabilities for each of the folds are then averaged for prediction. We train ``competence'' classifiers ($\mathcal{C}$) on the dataset $\mathcal{D}^E$ using logistic regression to predict ``competence'' scores between $0-1$ for base classifiers. If base classifier $B_i$ gets a ``competence'' score greater than $0.5$ then the base classifier is considered competent. 
To derive features from CNN, we use pre-trained models presented by the VGG-16 team in the ILSVRC-2014 competition \cite{DBLP:journals/corr/SimonyanZ14a} and the  CAFFE framework \cite{Jia:2014:CCA:2647868.2654889}. For deep tags, we consider top $d=10$ object labels as $d=10$ worked best.

\subsubsection*{\bf Exploratory Analysis.} We provide exploratory analysis in Table \ref{table:exp} to highlight the potential of merging object, scene and tag 
\begin{wraptable}{r}{0.29\textwidth}
\centering
\begin{small}
\begin{tabular}{l|c|c|c}
{Test} & {Pr(\%)} & {Pu(\%)} & {O(\%)} \\
\hline
Object is correct & 49 & 95.7 & 84.8\\
Scene is correct & 51 & 94.7 & 84.4\\
Tag is correct & 57 & 91.1 & 83 \\


All are correct & 30 & 87.3 & 73.9 \\ 

All are wrong & 27 & 1.5 & 7.4 \\

Atleast one  & {\color{blue} \bf 73} & {\color{blue} \bf 98.5} & {\color{blue} \bf 92.6} \\
modality is correct &&&\\

\end{tabular}
\end{small}
\caption{Exploratory analysis.} 
\label{table:exp}
\vspace{-8mm}
\end{wraptable}
modality for privacy prediction.  We predict privacy for images in the {\em Test} set using base classifiers in $\mathcal{B}$ and obtain  ``private'' (Pr), ``public'' (Pu) and ``overall'' (O) accuracy for: (a) a modality is correct (e.g., object), (b) all modalities are correct, (c) all modalities are wrong, and (d) at least one modality is correct. Table \ref{table:exp} shows that out of the three base classifiers (top 3 rows), the tag model yields the best accuracy for the private class (57\%). Interestingly,  the results for ``at least one modality is correct'' (73\%) show that using multi-modality, there is a huge potential (16\%) to improve the performance of the private class. This large gap is a promising result for developing multi-modality approaches for privacy prediction. Next, we evaluate DMFP that achieved the best boost in the performance for the private class using these modalities.

\subsection{Impact of Parameters $k_v$ and $k_p$ on DMFP}
\begin{wrapfigure}{r}{0.26\textwidth}
\centering
\includegraphics[scale=0.29]{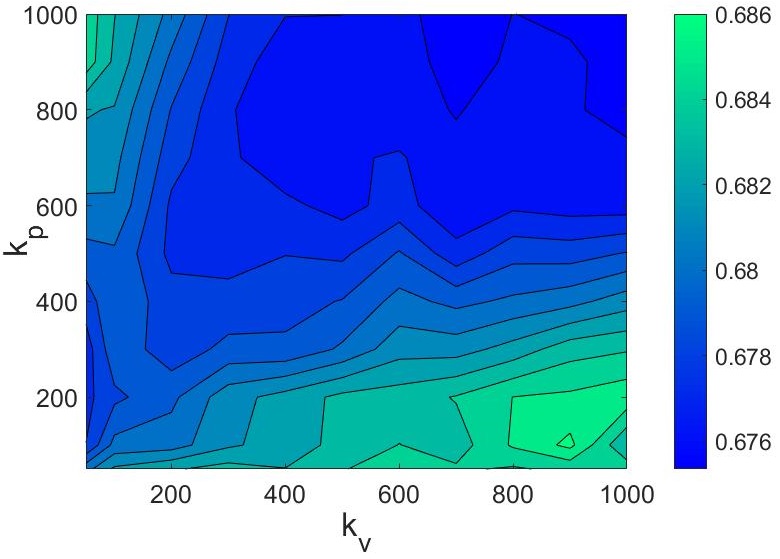}
\vspace{-5.2mm}
\caption{\label{fig:paramest} F1-measure obtained for various {\em $k_v$} and {\em $k_{p}$} values.}
\vspace{-3mm}
\end{wrapfigure}

We show the impact of neighborhood parameters, i.e., $k_v$ and $k_p$ on the privacy prediction performance obtained by the proposed approach DMFP. $k_v$ and $k_p$ are used to identify visual ($N_V$) and privacy profile ($N_P$) neighborhoods of a target image, respectively (Alg. \ref{classifierselection} lines 7--8). We experiment with a range of values for both the parameters $k_v, k_{p} = \{10, 20, \cdots, 100, 200, \cdots, 1000\}$, in steps of 
$10$ upto $100$ and then in steps of $100$. We also experiment with larger $k_v$ and $k_p$ values, but for better visualization, we only show the values with significant results. Figure \ref{fig:paramest} shows the F1-measure 
obtained (using 3-fold cross-validation on the $\mathcal{D}^E$ dataset) for the private class for various $k_v$ and $k_p$ values. We notice that when we increase the $k_v$ parameter the performance increase whereas when we increase $k_p$ parameter, the performance increases upto $k_p = 200$, then the performance decreases gradually. The results show that the performance is quite sensitive to changes in the privacy neighborhood ($N_P$) parameter $k_p$, but relatively insensitive to changes in the visual neighborhood ($N_V$) parameter $k_v$.
We get the best performance for $k_v = 900$ and $k_{p} = 100$. We use these parameter values in the next experiments.

\begin{table*}[t]
\renewcommand{\arraystretch}{0.95}
\centering
\begin{small}
\begin{tabular}{|l|c|c|c|c|c|c|c|c|c|c|}
\hline
& \multicolumn{3}{|c|}{Private} & \multicolumn{3}{|c|}{Public} & \multicolumn{4}{|c|}{Overall}\\
{Features} & {Precision} & {Recall} & {F1-score} & {Precision} & {Recall} & {F1-score} & {Accuracy (\%)} & {Precision} & {Recall} & {F1-score} \\
\hline
DMFP &  0.752 & {\color{blue}{\bf 0.627}} & {\color{blue}{\bf 0.684}} & {\color{blue}{\bf 0.891}} & 0.936 & {\color{blue}{\bf 0.913}} & {\color{blue}{\bf 86.36}} & {\color{blue}{\bf 0.856}} & {\color{blue}{\bf 0.859}} & {\color{blue}{\bf 0.856}}\\
\hline
\multicolumn{11}{|c|}{Components of the proposed approach}\\
\hline
DMFP$-{N_V}$  & 0.763 & 0.575 & 0.655 & 0.879 & 0.945 & 0.91 & 85.79 & 0.85 & 0.852 & 0.847\\
DMFP$-{N_P}$  & 0.74 & 0.572 & 0.645  & 0.877 & 0.938  & 0.907 & 85.21  & 0.843 & 0.847 & 0.841\\
${N_V}-CL$  & {\color{blue}{\bf 0.79}} & 0.534 & 0.637 & 0.87 & {\color{blue}{\bf 0.956}} & 0.911 & 85.71 & 0.85 & 0.851 & 0.843 \\

${N_P}-CL$ & 0.788 & 0.537 & 0.639 & 0.87 & {\color{blue}{\bf 0.956}} & 0.911 & 85.73 & 0.85 & 0.851 & 0.843 \\


$\{N_V + N_P\}-CL$ & {\color{blue}{\bf 0.79}} & 0.534 & 0.637 & 0.87 & {\color{blue}{\bf 0.956}} & 0.911 & 85.71 & 0.85 & 0.851 & 0.843 \\

\hline
\multicolumn{11}{|c|}{``Competence'' Features}\\
\hline






DMFP$-\phi_1$ & 0.777 & 0.553 & 0.646 & 0.874 & 0.951 & 0.911 & 85.74 & 0.849 & 0.852 & 0.844 \\


DMFP$-\phi_2$ & 0.74 & 0.565 & 0.641 & 0.875 & 0.939 & 0.906 & 85.11 & 0.842 & 0.846 & 0.84 \\


DMFP$-\phi_3$ & 0.752 & 0.627 & 0.683 & 0.891 & 0.936 & 0.913 & 86.35 & 0.856 & 0.859 & 0.856 \\


\hline
\end{tabular}
\end{small}
\caption{Evaluation of dynamic multi-modal fusion for privacy prediction (DMFP). } 
\label{table:algeval}
\end{table*}

\subsection{Evaluation of the Proposed Approach}

We evaluate the proposed approach DMFP for privacy prediction in an ablation experiment setting. Specifically, we remove a particular component of the proposed approach DMFP and compare the performance of DMFP before and after the removal of that component. We consider excluding several components from DMFP: (1) the visual neighborhood $N_V$ (DMFP$-N_V$), (2) the privacy profile neighborhood $N_P$ (DMFP$-N_P$), (3) ``competence'' features (e.g., DMFP$ - \phi_1$), and (4) base classifier selection without ``competence'' learning (e.g., $N_V-CL$). For option (4), we consider a simpler version of the proposed algorithm, in which we do not learn a competence classifier for a base classifier; instead, we rely solely on the number of accurate predictions of the samples from a neighborhood. We evaluate it using images from three regions: (a) neighborhood $N_V$ only ($N_V-CL$), (b) neighborhood $N_P$ only ($N_P-CL$), and (c) both the neighborhoods $N_P$ and $N_V$ ($\{N_P + N_V\}-CL$).  


Table \ref{table:algeval} shows the class-specific (private and public) and overall performance obtained by the proposed approach (DMFP) and after removal of its various components detailed above. Primarily, we wish to identify whether the proposed approach characterizes the private class effectively as sharing private images on the Web with everyone is not desirable. 
We observe that the proposed approach achieves the highest recall of $0.627$ and F1-score of $0.684$ (private class), which is better than the performance obtained by eliminating the essential components (e.g., neighborhoods) of the proposed approach. We notice that if we remove either of the neighborhood $N_V$ or $N_P$, the recall and F1-score drop by $5\%$ and $4\%$. This suggests that both the neighborhoods ($N_V$, $N_P$) are required to identify an appropriate local region surrounding a target image. It is also interesting to note that the performance of DMFP$-N_P$ (removal of $N_P$) is slightly lower than the performance of DMFP$-N_V$ (removal of $N_V$), depicting that neighborhood $N_P$ is helping more to identify the competent base classifier(s) for a target image. The $N_P$ neighborhood brings images closer based on their privacy probabilities and hence, is useful to identify the competent base classifier(s) (this is evident in Figure \ref{fig:AlgIll}). 
We also show that when we remove competence learning (CL) i.e., $N_V-CL$, $N_P-CL$, and $\{NV + NP\}-CL$, the precision improves by $4\%$ (private class), but the recall and F1-score (private class) drops by $9\%$ and $5\%$ respectively, showing that competence learning is necessary to achieve the best performance.

We also remove the ``competence'' features one by one and record the performance of DMFP to understand which competence features are essential. Table \ref{table:algeval} shows that when we remove feature $\phi_1$ corresponding to the neighborhood $N_V$, the performance drops significantly ($\approx 4\%$). 
Likewise, when we remove $\phi_2$ (feature corresponding the $N_P$ region), we notice a similar decrease of $4\%$ in the F1-score of private class. Note that, when we remove the ``competence'' features corresponding to their neighborhoods (such as $\phi_1$ for $N_V$ and $\phi_2$ for $N_P$), we get nearly similar performance as we remove the respective neighborhoods from the proposed approach (DMFP$-N_V$ and DMFP$-N_P$); implying that removing ``competence'' features (e.g., $\phi_1$) is as good as removing the corresponding neighborhood ($N_V$). However, a close look at the performance suggests that the performance obtained using DMFP$-\phi_1$ (recall of $0.553$) is slightly worse than the performance of DMFP$-N_V$ (recall of $0.57$). Similarly, for DMFP$-\phi_2$, the performance (recall) decrease from $0.572$ obtained using DMFP$-N_P$ to $0.565$. The performance decrease can be explained as when we remove the neighborhood $N_V$ or $N_P$, the respective ``competence'' features are empty, and that might be helpful for some cases (as zero-valued $\phi_2$ feature of $\phi^s$ was helpful in Figure \ref{fig:AlgIll}). Additionally, the recall of DMFP$-N_V$ and DMFP$-N_P$ are similar whereas the recall of DMFP$-\phi_1$ ($0.553$) is slightly worse than the recall of DMFP$-\phi_2$ ($0.565$). The results suggests that the neighborhood $N_V$ is more dependent on the ``competence'' features as compared to the neighborhood $N_P$. We experimented with the probability based ``competence'' features (instead of boolean features), but did not yield improvements in the performance.

\begin{table*}[t]
\centering
\begin{small}
\begin{tabular}{|l|c|c|c|c|c|c|c|c|c|c|}
\hline
& \multicolumn{3}{|c|}{Private} & \multicolumn{3}{|c|}{Public} & \multicolumn{4}{|c|}{Overall}\\
{Features} & {Precision} & {Recall} & {F1-score} & {Precision} & {Recall} & {F1-score} & {Accuracy (\%)} & {Precision} & {Recall} & {F1-score} \\
\hline
\hline
DMFP &  0.752 & {\color{blue}{\bf 0.627}} & {\color{blue}{\bf 0.684}} & {\color{blue}{\bf 0.891}} & 0.936 & {\color{blue}{\bf 0.913}} & {\color{blue}{\bf 86.36}} & {\color{blue}{\bf 0.856}} & {\color{blue}{\bf 0.859}} & {\color{blue}{\bf 0.856}}\\
\hline
\hline
Object ($B^o$) & {\color{blue}{\bf 0.772}} & 0.513 & 0.616 & 0.864 & {\color{blue}{\bf 0.953}} & 0.907 & 84.99 & 0.843 & 0.85 & 0.838 \\
Scene ($B^s$) & 0.749 & 0.51 & 0.606 & 0.863 & 0.947 & 0.903 & 84.45 & 0.836 & 0.844 & 0.833 \\
Image tags ($B^t$) & 0.662 & 0.57 & 0.612 & 0.873 & 0.91 & 0.891 & 83.03 & 0.823 & 0.83 & 0.826 \\

\hline
\end{tabular}
\end{small}
\caption{Dynamic multi-modal fusion for privacy prediction (DMFP) vs. base classifiers of DMFP.} 
\label{table:featureset}
\end{table*}

\begin{table*}[]
\centering
\begin{small}
\begin{tabular}{|l|c|c|c|c|c|c|c|c|c|c|}
\hline
& \multicolumn{3}{|c|}{Private} & \multicolumn{3}{|c|}{Public} & \multicolumn{4}{|c|}{Overall}\\
{Features} & {Precision} & {Recall} & {F1-score} & {Precision} & {Recall} & {F1-score} & {Accuracy (\%)} & {Precision} & {Recall} & {F1-score} \\
\hline
\hline
DMFP &  0.752 & {\color{blue}{\bf 0.627}} & {\color{blue}{\bf 0.684}} & {\color{blue}{\bf 0.891}} & 0.936 & {\color{blue}{\bf 0.913}} & {\color{blue}{\bf 86.36}} & {\color{blue}{\bf 0.856}} & {\color{blue}{\bf 0.859}} & {\color{blue}{\bf 0.856}}\\
\hline
\hline




\citet{DBLP:conf/aaai/ZahavyKMM18} & 0.662 & 0.568 & 0.612 & 0.873 & 0.911 & 0.891 & 83.02 & 0.82 & 0.825 & 0.821 \\

Majority Vote & {\color{blue}{\bf 0.79}} & 0.534 & 0.637 & 0.87 & {\color{blue}{\bf 0.956}}& 0.911  & 85.71 & 0.85 & 0.851 & 0.843 \\

Decision-level Fusion & 0.784 & 0.555 & 0.65 & 0.874 & 0.953 & 0.912 & 85.94  & 0.852 & 0.853 & 0.846 \\

Stacked--En & 0.681 & 0.59 & 0.632 & 0.879 & 0.915 & 0.897 & 83.86 & 0.829 & 0.834 & 0.831 \\

Cluster--En & 0.748 & 0.429 & 0.545 & 0.845 & {\color{blue}{\bf 0.956}}
& 0.897 & 83.17 & 0.822 & 0.831 & 0.814 \\

\hline
\end{tabular}
\end{small}
\caption{Dynamic multi-modal fusion for privacy prediction (DMFP) vs. baselines.} 
\label{table:baselinecomp}
\end{table*}

\subsection{Proposed Approach vs. Base Classifiers} \label{exp:featuresets}
We compare privacy prediction performance obtained by the proposed approach DMFP with the set of base classifiers $\mathcal{B}$: 1. object ($B^o$), 2. scene ($B^s$), and 3. image tags ($B^t$). 



Table \ref{table:featureset} compares the performance obtained by the proposed approach (DMFP) and base classifiers. We achieve the highest performance as compared to the base classifiers and show a maximum improvement of $\approx 10\%$ in the F1-score of private class. We notice that our approach based on multi-modality yields an improvement of $11\%$ over the recall of object and scene models and an improvement of $\approx 6\%$ over the recall of the tag model, that is the best-performing single modality model obtained for the private class from the exploratory analysis (refer to Table \ref{table:exp}). Still, our approach makes some errors (See Table \ref{table:exp} and \ref{table:algeval}, $73\%$ vs. $62\%$). A close look at the errors discovered that a slight subjectivity of annotators could obtain different labels for similar image subjects (e.g., food images are very subjective). 

\subsubsection*{\bf Error Analysis}

\begin{figure}[t]
\renewcommand{\arraystretch}{0.9}
\centering
\begin{small}
\begin{tabular}{|c|cccc|}
\hline

Model &
\includegraphics[scale=0.07]{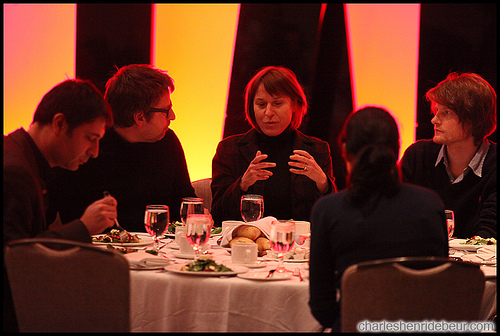} & \includegraphics[scale=0.06]{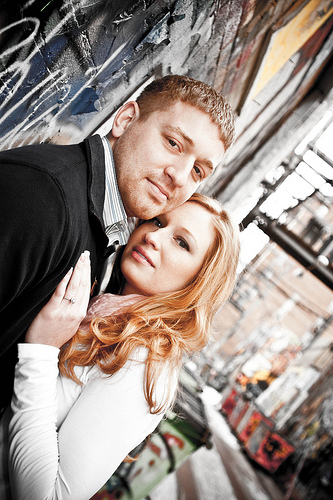} & \includegraphics[scale=0.06]{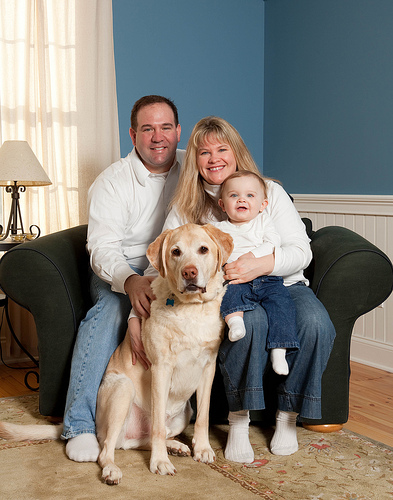} & \includegraphics[scale=0.05]{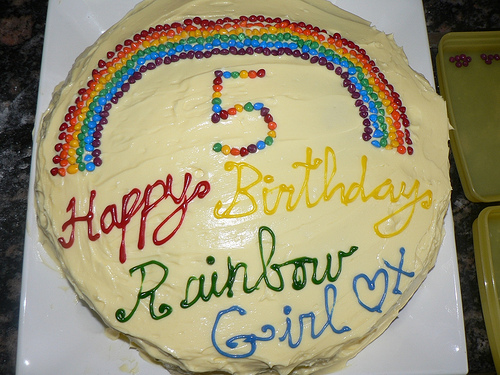}\\

& (a) & (b) & (c) & (d)\\

\hline

DMFP & \cmark & \cmark & \cmark & \xmark\\
Object & \xmark & \cmark & \cmark & \xmark\\
Scene & \cmark & \xmark & \cmark & \xmark\\
Tags & \cmark & \cmark & \xmark & \xmark\\


\hline
\end{tabular}	 
\end{small}
\caption{Predictions for private images.} 
\label{fig:results}
\end{figure}

We perform error analysis to further analyze 
\begin{wraptable}{r}{0.26\textwidth}
\centering
\begin{small}
\begin{tabular}{l|c|c|c}
\hline
{$\mathcal{B}$} & {overall} & {private} & {public}  \\
\hline
object & 16.52 & 14.00 & 23.75 \\
scene & 27.71 & 21.78 & 42.63  \\
Tags & 37.02 & 26.90 & 58.79 \\
\hline
\end{tabular}
\end{small}
\caption{Errors corrected (\%).} 
\label{table:err}
\end{wraptable}
the results of the proposed approach. We first determine the errors generated by all the base classifiers in $\mathcal{B}$ and corrected by the proposed approach DMFP. We calculate the percentage of corrected errors for private class, public class and overall (considering both the classes) and show them in Table \ref{table:err}. We compute the percentage of corrected errors as the number of corrected errors of private (or public) class over the total number of private (or public) class errors. We calculate the fraction of overall corrected errors by considering both public and private classes. The table shows that we correct $14\% - 27\%$ of private class errors, $18\% - 58\%$ of public class errors and overall we eliminate $16\% - 37\%$ errors. Note that errors generated for the private class are much larger than the public class (See Table \ref{table:featureset}) and hence, even a comparatively smaller percentage of corrected errors constitute to a significant improvement. We also analyze results by showing predictions of samples in Figure \ref{fig:results}, for which at least one base classifier fails to predict the correct privacy of an image.  For instance, for example (b), scene model failed to predict the correct privacy of the image; however, DMFP identifies the competent base classifiers, i.e., object, and tag and predict the correct privacy label. We also show an example (image (d)) for which all the base classifiers fail to predict the correct privacy class and hence, the proposed approach also fails to predict the correct label. The image of a food is very subjective and hence, generic base classifiers will not be sufficient to predict the correct labels of such images. In the future, these generic models can be extended to develop hybrid approaches, that consider both generic and subjective privacy notions to predict  personalized privacy labels.

\begin{table*}[t]
\centering
\begin{small}
\begin{tabular}{|l|c|c|c|c|c|c|c|c|c|c|}
\hline
& \multicolumn{3}{|c|}{Private} & \multicolumn{3}{|c|}{Public} & \multicolumn{4}{|c|}{Overall}\\
{Features} & {Precision} & {Recall} & {F1-score} & {Precision} & {Recall} & {F1-score} & {Accuracy (\%)} & {Precision} & {Recall} & {F1-score} \\
\hline
\hline
DMFP &  {\color{blue}{\bf 0.752}} & {\color{blue}{\bf 0.627}} & {\color{blue}{\bf 0.684}} & {\color{blue}{\bf 0.891}} & {\color{blue}{\bf 0.936}} & {\color{blue}{\bf 0.913}} & {\color{blue}{\bf 86.36}} & {\color{blue}{\bf 0.856}} & {\color{blue}{\bf 0.859}} & {\color{blue}{\bf 0.856}}\\
\hline
\hline
PCNH \cite{Tran:2016:PFD:3015812.3016006} & 0.689 & 0.514 & 0.589 & 0.862 & 0.929 & 0.894 & 83.15 & 0.819 & 0.825 & 0.818 \\
Concat \cite{DBLP:conf/aaai/TongeC18} ($F^{ost}$) & 0.671 & 0.551 & 0.605 & 0.869 & 0.917 & 0.892 & 83.09 & 0.82 & 0.826 & 0.821 \\

\hline
\end{tabular}
\end{small}
\caption{Dynamic multi-modal fusion for privacy prediction (DMFP) vs. prior image privacy prediction works.} 
\label{table:priorworkcomp}
\end{table*}

\subsection{Proposed Approach vs. Baselines}


We compare the performance of the proposed approach DMFP with multi-modality based baselines described below. 


{\bf 1. Model Selection by \citet{DBLP:conf/aaai/ZahavyKMM18}:} 
The authors proposed a deep multi-modal architecture for product classification in e-commerce, wherein they learn a decision-level fusion policy to choose between image and text CNN for an input product. Specifically, the authors provide class probabilities of a product as input to the policy trained on a validation dataset and use it to predict whether image CNN (or text CNN) should be selected for the input. In other words, policy determines the competence of the CNNs for its input and thus, we consider it as our baseline. For a fair comparison, we learn $3$ policies (corresponding to the competence classifiers $\mathcal{C}$), wherein each policy (say object policy) predicts whether the respective base classifier (object) should be selected for a target image. Note that we learn these policies on the $\mathcal{D}^E$ dataset. Finally, we take a majority vote of the privacy label predicted by the selected base classifiers (identified by the policies) for a target image.

{\bf 2. Majority Vote:} We consider a majority vote as another baseline, as we use it for final selection of privacy label for a target image. Unlike our approach, a vote is taken without any pre-selection of base classifiers. We predict privacy of a target image using base classifiers in $\mathcal{B}$ and select a label having highest number of votes.

{\bf 3. Decision-level Fusion:} Fixed rules, that average the predictions of the different CNNs \cite{NIPS2012_4824} or select the CNN with the highest confidence \cite{Poria:2016:FAV:2852370.2852645}. The first rule is equivalent to the majority vote baseline, and hence, we show the results for the second rule only. The second rule is given as: $Y_T = argmax_i([{P_i}^o + {P_i}^s + {P_i}^t] / 3)$, where $i = 0$ (public), $1$ (private). $P^o, P^s,$ and $P^t$ denotes the posterior probabilities (private or public) obtained using object ($B^o$), scene ($B^s$) and tag ($B^t$) modality respectively.

{\bf 4. Stacked Ensemble (Stacked--en): }  
Stacking learns a meta-classifier to find an optimal combination of the base learners \cite{superlearner,DBLP:journals/ml/Breiman96a}. Unlike bagging and boosting, stacking ensembles robust and diverse set of base classifiers together, and hence, we consider it as one of the baselines. We use the same set of base classifiers $\mathcal{B}$ to encode images in $\mathcal{D}^T$ using posterior probabilities $P(Y_I = private| I, B_i)$ and $P(Y_I = public| I, B_i)$ where $B_i \in \mathcal{B}$. We train a meta-classifier on this encoded ${\mathcal{D}^T}$ dataset using calibrated SVM classifier. We use this meta-classifier to predict privacy class of an encoded target image $T$ (using the posterior probabilities obtained by the base classifiers $P(Y_T=private | T, B_i)$ and $P(Y_T=public | T, B_i)$). As we use ${\mathcal{D}^E}$ only to learn ``competence'' classifiers, we do not consider it for training a meta-classifier for a fair comparison.  

{\bf 5. Clusters-based Models (Cluster--en): }  We create $5$ clusters of ${\mathcal{D}^T}$ dataset using hierarchical clustering mechanism and the combination of object, scene and tag features $(F^{ost})$. We train a calibrated SVM model on each cluster using the combination of features $F^{ost}$. For target image $T$, the most relevant cluster is identified using $k = 15$ nearest neighbors, and the model trained on that cluster is used to predict the privacy of the target image. We consider this as another baseline, because clustering images that are shared online, brings similar image types (e.g., portraits) together and models trained on these clusters can be competent to predict privacy of target images of respective image types. The number of clusters and neighbors are estimated based on the ${\mathcal{D}^E}$ dataset.

Table \ref{table:baselinecomp} compares the performance obtained by the proposed approach DMFP with the performance obtained using the baseline models. We observe that DMFP learns better privacy characteristics than baselines with respect to private class by providing improvements of $4.5\%-14\%$ and $4\%-20\%$ in the F1-score and recall of private class. 
When we learn the ``competence'' of the base classifiers ($\mathcal{B}$) on the $\mathcal{D}^E$ dataset without identifying the neighborhoods (the first baseline, \citet{DBLP:conf/aaai/ZahavyKMM18}), the precision, recall and F1-score drop by $9\%$, $\approx 6\%$, $\approx 7\%$. It is interesting to note that the precision of DMFP$ - CL$ (Refer Table \ref{table:algeval}, $N_V-CL$, $N_P-CL$, $\{N_V + N_P\}-CL$), i.e., $0.79$ is better than the first baseline (\citet{DBLP:conf/aaai/ZahavyKMM18}), i.e., $0.662$ whereas the recall of the first baseline ($0.568$) is better than DMFP$-CL$ ($0.534$). However, when we combine the neighborhoods ($\{N_V + N_P\}$) and the first baseline (competence learning), i.e., the proposed approach DMFP, we get better performance than each of these methods. Another detail to note that the performance of the first baseline (\citet{DBLP:conf/aaai/ZahavyKMM18}) is very close to the image tags model (see Table \ref{table:baselinecomp}, \ref{table:featureset}) and even though the baseline uses multi-modality, the performance does not exceed significantly over the individual base classifiers (object, scene, image). \citet{DBLP:conf/aaai/ZahavyKMM18} performed well for product classification, but it failed to yield improved results for privacy prediction because unlike product images or ImageNet images (that contains single object in the image), images that are shared online are much more complex (containing multiple objects, and scene) and diverse (having different image subjects such as self-portraits, personal events). The results suggest that it is hard to generalize the competency of base classifiers on all types of image subjects and hence, the competence of the base classifiers needs to be determined dynamically based on the neighborhoods of a target image.

Table \ref{table:baselinecomp} also shows that F1-measure of private class improves from $0.637$ achieved by majority vote (the second baseline), $0.65$ obtained by decision-level fusion (the third baseline), $0.636$ obtained by stacked--en (the fourth baseline), and $0.545$ obtained by cluster--en (the fifth baseline) to $0.684$ obtained by DMFP. Additionally, we notice that the proposed approach is able to achieve a performance higher than $85\%$ in terms of all compared measures. Note that a naive baseline which classifies every image as ``public'' obtains an accuracy of $75\%$. With a paired T-test, the improvements over the baseline approaches for F1-measure  of a private class are statistically significant for p-values $< 0.05$.


\subsection{Proposed Approach vs. Prior Image Privacy Prediction Works}
We compare the privacy prediction performance obtained by the proposed approach DMFP with the state-of-the-art works of privacy prediction: {\bf 1. object} \cite{tongemsm18, DBLP:conf/aaai/TongeC16} ($B^o$), {\bf 2. scene} \cite{DBLP:conf/aaai/TongeC18} ($B^s$), {\bf 3. image tags} \cite{tongemsm18, DBLP:conf/aaai/TongeC16,Squicciarini2014} ($B^t$), 4.  PCNH privacy framework \cite{Tran:2016:PFD:3015812.3016006}, and 5. Concatenation of all features \cite{DBLP:conf/aaai/TongeC18}. Note that the first three works are the feature sets of DMFP and are evaluated in the Experiment \ref{exp:featuresets}.  
We describe the remaining prior works (i.e., 4 and 5) in what follows. 
{\bf 4. PCNH privacy framework} \cite{Tran:2016:PFD:3015812.3016006}:
The framework combines features obtained from two architectures: one that extracts convolutional features (size = $24$), and another that extracts object features (size = $24$). The object CNN is a very deep network of $11$ layers obtained by appending three fully-connected layers of size $512$, $512$, $24$ at the end of the fully-connected layer of AlexNet \cite{NIPS2012_4824}. The PCNH framework is first trained on the ImageNet dataset \cite{ILSVRC15} and then fine-tuned on a privacy dataset.
{\bf 5. Combination of Object, Scene and User Tags (Concat)} \cite{DBLP:conf/aaai/TongeC18}: 
\citet{DBLP:conf/aaai/TongeC18} combined object and scene tags with user tags and achieved an improved performance over the individual sets of tags. Thus, we compare the proposed approach with the SVM models trained on the combination of all feature sets ($F^{ost}$) to show that it will not be adequate to predict the privacy of an image accurately. In our case, we consider object and scene visual features instead of tags and combine them with user tags to study multi-modality with the concatenation of visual and tag features.

Table \ref{table:priorworkcomp} compares the performance obtained by the proposed approach (DMFP) and prior works. We achieve the highest performance as compared to the prior works and show a maximum improvement of $\approx 10\%$ in the F1-score of private class. We notice that our approach based on multi-modality yields an improvement of $11\%$ over the recall of almost all the prior works (Refer Table \ref{table:featureset} and \ref{table:priorworkcomp}). 
Particularly, we show improvements in terms of all measures over the PCNH framework, that uses two kinds of features object and convolutional. We found that adding high-level descriptive features such as scene context and image tags to the object features help improve the performance. In addition to the individual feature sets, we also outperform the concatenation of these feature sets (denoted as ``Concat''), showing that ``Concat'' could not yield an optimal combination of multi-modality.  We notice that the performance of ``Concat'' is slightly lower than the performance of base classifiers (Refer Tables \ref{table:featureset} and \ref{table:priorworkcomp}). We find this is consistent with \citet{DBLP:conf/aaai/ZahavyKMM18} results, that concatenated various layers of image and tag CNN and trained the fused CNN end-to-end but did not yield a better performance than the individual CNN (image or tag). 

\section{Conclusions and Future Work}
In this work, we estimate the competence of object, scene and image tag modalities, derived through convolutional neural networks and dynamically identify the set of most competent modalities for a target image to adequately predict the class of the image as {\em private} or {\em public}.  
The proposed approach contains three stages wherein we first identify neighborhoods for a target image based on visual content similarity and privacy profile similarity.  Then, we derive ``competence'' features from these neighborhoods and provide them to the ``competence'' classifiers to predict whether a modality is competent for the target image. Lastly, we select the subset of the most competent modalities and take a majority vote to predict privacy class of the target image.
Experimental results show that our approach predicts the sensitive (or private) content more accurately than the models trained on an individual modality (object, scene, and tags), multi-modality baselines and prior privacy prediction approaches.
Also, our approach could aid other applications such as event understanding, image classification, to on the fly decide which CNN (object, scene or tag) to use based on a target image.

In the future, it will be interesting to study dynamic multi-modal fusion in personalized privacy setting. Also, other types of competence learning approaches 
and competence features can be developed for estimating the competence of base classifiers. 

\section*{Acknowledgments}
This research is supported by the NSF grant \#1421970. The computing for this project was performed on Amazon Web Services.

\balance

\bibliographystyle{ACM-Reference-Format}
\bibliography{ref}

\end{document}